\def\year{2020}\relax
\documentclass[letterpaper]{article} 
\usepackage{aaai20}  
\usepackage{times}  
\usepackage{helvet} 
\usepackage{courier}  
\usepackage[hyphens]{url}  
\usepackage{graphicx} 
\usepackage{graphicx}  
\usepackage{booktabs}
\usepackage{amsmath}
\newcommand{\citet}[1]{\citeauthor{#1} \shortcite{#1}} 
\newcommand{\citep}{\cite}

\title{Posterior-GAN: Towards Informative and Coherent Response Generation with Posterior Generative Adversarial Network}
\author{
	Shaoxiong Feng,\textsuperscript{\rm 1}\thanks{Work done at Data Science Lab, JD.com.}
	Hongshen Chen,\textsuperscript{\rm 2}
	Kan Li,\textsuperscript{\rm 1}
	Dawei Yin\textsuperscript{\rm 2} \\ 
	\textsuperscript{\rm 1} School of Computer Science \& Technology, Beijing Institute of Technology \\
	\textsuperscript{\rm 2} Data Science Lab, JD.com \\
	\{shaoxiongfeng,likan\}@bit.edu.cn, \\
    ac@chenhongshen.com,
	yindawei@acm.org 
}

\begin{document}
\maketitle

\begin{abstract}
Neural conversational models learn to generate responses by taking into account the dialog history. These models are typically optimized over the \textit{query-response} pairs with a maximum likelihood estimation objective. However, the query-response tuples are naturally loosely coupled, and there exist multiple responses that can respond to a given query, which leads the conversational model learning burdensome. Besides, the general dull response problem is even worsened when the model is confronted with meaningless response training instances. Intuitively, a high-quality response not only responds to the given query but also links up to the future conversations, in this paper, we leverage the \textit{query-response-future turn} triples to induce the generated responses that consider both the given context and the future conversations. To facilitate the modeling of these triples, we further propose a novel encoder-decoder based generative adversarial learning framework, Posterior Generative Adversarial Network (Posterior-GAN), which consists of a forward and a backward generative discriminator to cooperatively encourage the generated response to be informative and coherent by two complementary assessment perspectives. Experimental results demonstrate that our method effectively boosts the informativeness and coherence of the generated response on both automatic and human evaluation, which verifies the advantages of considering two assessment perspectives.
\end{abstract}

\section{Introduction}

Generative conversational models are drawing an increasing amount of interests \cite{shang2015neural,vinyals2015neural,serban2016building,serban2017hierarchical,serban2017multiresolution,li2016diversity,li2016deep,li2017adversarial,mou2016sequence,zhao2017learning,xing2017topic,xu2017neural,xu2018diversity,zhang2018reinforcing,zhang2018generating}. Most existing generative conversational models are based on a Seq2Seq architecture \cite{sutskever2014sequence}. These models consider conversation history to learn to generate responses and are optimized over the \textit{query-response} pairs.

\begin{figure}
    \centering
    \includegraphics[width=.95\columnwidth]{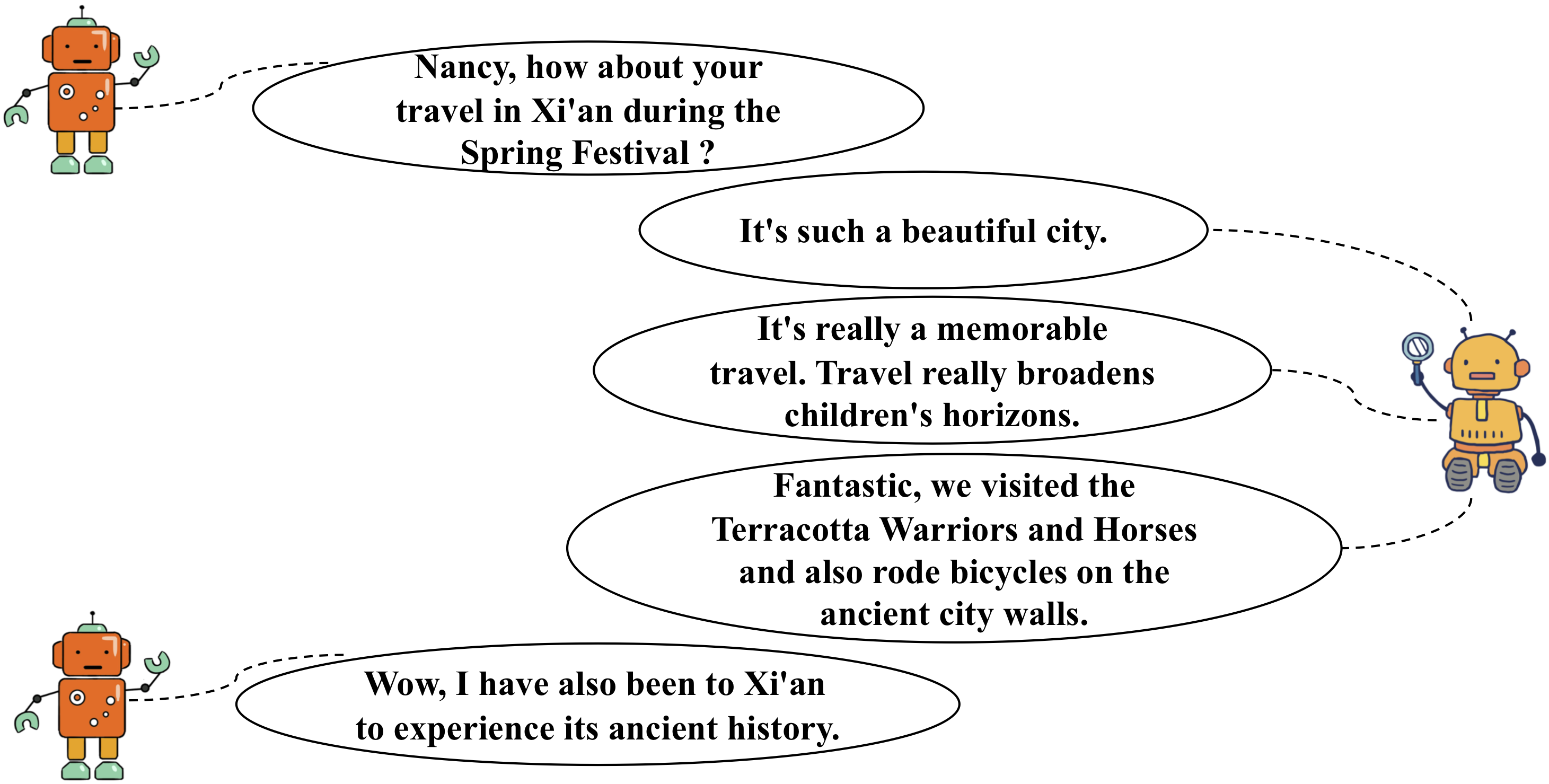}
    \caption{Dialogue examples with various responses regarding informativeness and coherence.}
    \label{fig:introduction_sample}
\end{figure}

However, the query-response tuples are naturally loosely coupled, there exist multiple responses that can respond to a given query, so call \textit{one-to-many} phenomenon, which leads the conversational model learning burdensome. In DailyDialog \cite{li2017dailydialog} corpus, at least 13\% utterances contain more than one response \cite{csaky2019improving}. What's more, the notorious general dull response problem is even worsened when the model is confronted with meaningless response training instances. In another public available corpus OpenSubtitles (OSDb), 113K sentences contain the sequence ``I don't know''  in the training set \cite{li2016diversity}. Not to mention other similar meaningless responses like ``haha", ``what are you talking about?", etc. The \textit{one-to-many} phenomenon and non-negligible proportion of generic responses in the training corpus cause the neural response generation model prone to generate short, bland, or even irrelevant responses. In Figure \ref{fig:introduction_sample}, for the given query talking about the traveling experience, the first response is much shorter and uninformative compared with the other two responses, while the second one seems to be informative enough but meanders away from the conversational subject, especially in terms of the following conversations. The third response is not only informative but also coherent with both the query and the next utterance.

It is often the case that a high-quality response not only responds to the given query but also links up to the future conversations, in this paper, we propose to utilize \textit{query-response-future turn} triples instead of the \textit{query-response} pairs to train the response generation model. Conventionally, neural dialogue generation model is optimized with a maximum likelihood estimation (MLE) objective given the \textit{query-response} tuples. However, such an objective is obviously inadequate for triple learning, where the future conversation is introduced during training. Moreover, the MLE objective encourages the model to repetitively overproduce high-frequency words in the ambiguous and noisy training corpus \cite{zhang2018reinforcing} and tend to deterministically output some ``average'' of diverse real-world responses \cite{csaky2019improving}. To extend the neural dialogue learning from tuples to triples, and induce the generated response to be not only informative but also coherent regarding both the input context and the future conversations, we further propose a novel encoder-decoder based generative adversarial learning framework, Posterior Generative Adversarial Network (Posterior-GAN), to handle the \textit{query-response-future turn} triple modeling. The framework leverages a forward and a backward generative discriminator to guide the generated response: the forward discriminator that extracts as much sentence-level semantic information as possible from the response to predict the real-world future conversations outputs high rewards if the generated response is informative enough with respect to the subsequent future conversations, and the backward discriminator that assesses the response based on the full information of real-world future conversations instead of the input context encourages the generated response to be more coherent in terms of the following conversations and guides the conversation smoothly linking up to the future turn. 

We highlight our contributions as follows:
\begin{itemize}
\item We identify an unexplored type of metadata, \textit{query-response-future turn} triples, for response generation. Compared to general \textit{query-response} tuples, the triples help the model use bidirectional information to learn the response generation in training.

\item We propose a novel encoder-decoder based generative adversarial learning framework, Posterior-GAN, to facilitate the \textit{query-response-future turn} modeling, which induces the generated response to be informative and coherent by constructing two generative discriminators, a forward one and a backward one respectively. 

\item We perform detailed experiments to demonstrate the effectiveness of the proposed framework and verifies the ability of bidirectional generative discriminators on assessing the quality of response.
\end{itemize}

\section{Method}

\subsection{Overview}
In this paper, we extend the conventional \textit{query-response} tuple $(x,y)$ neural dialogue learning into \textit{query-response-future turn} triple $(x,y,z)$ to encourage the generated response to be informative and coherent with respect to both the given query and the future conversations.
 
\begin{figure}
    \centering 
    \includegraphics[width=.95\columnwidth]{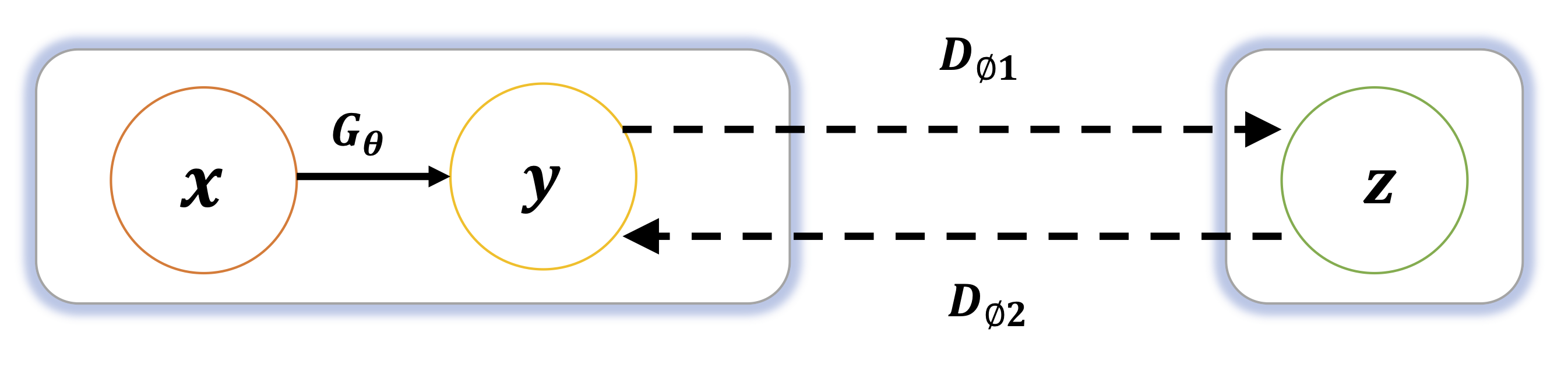}
    \caption{Illustration of Posterior-GAN. Brown for the query, yellow for the current response, and green for the future turn. $G_{\theta}$ represent generator. $D_{\phi 1}$ and $D_{\phi 2}$ represent forward and backward generative discriminator respectively.}
    \label{fig:theArchitectureOfModel}
\end{figure}

Here, a novel posterior generative adversarial network (Posterior-GAN) is proposed to undertake the triple learning and mitigate the overproduction of repetitive responses problem under the MLE objective. Posterior-GAN contains a generator that is responsible for generating response, and two discriminators cooperatively discriminating whether the generated response is coherent and informative in a forward and a backward manner by taking both the preceding context and the future conversations into account. The generator $G_{\theta}$ is constructed upon the Seq2Seq structure. Given the input query $x=\left(x_{1}, \dots, x_{t}, \ldots, x_{T}\right)$ of $T$ words from the vocabulary $\Gamma$, the model generates response $y=\left(y_{1}, \ldots,  y_{m}, \dots,  y_{L}\right)$ of $L$ words. For the discriminator, instead of a traditional classification-based discriminator, we utilize two symmetric generative discriminators with cross-entropy based rewards: a forward generative discriminator $D_{\phi 1}$ and a backward generative discriminator $D_{\phi 2}$. The general architecture is illustrated in Figure \ref{fig:theArchitectureOfModel}.

\subsection{Generator}
In our implementation, the generator consists of a two-layer bidirectional LSTM encoder and a four-layer LSTM decoder. The word embedding is sequentially fed to two-layer bidirectional LSTM  resulting with a hidden state representing the past and future information simultaneously. To better handle the long-range dependencies in multi-turn conversations, we also apply attention mechanism \cite{bahdanau2014neural} in the decoding phase.

\subsection{Discriminator}
Traditional discriminators in generative adversarial networks are classification-based approach, such as a binary classifier, which takes in the query-response pair $(x, y)$ and recognizes the true probability of pair $(x, y)$ being true as a reward. Essentially, it models the joint probability $p(x,y)$. However, as \citet{xu2018diversity} illustrated, when a query-generated response pair fits the distribution of real-world pairs, the classifier-based discriminators may result in saturated similar indistinguishable rewards for both the synthesised response and the ground truth response. 
 
\begin{figure} [!t]
    \centering 
    \includegraphics[width=.95\columnwidth]{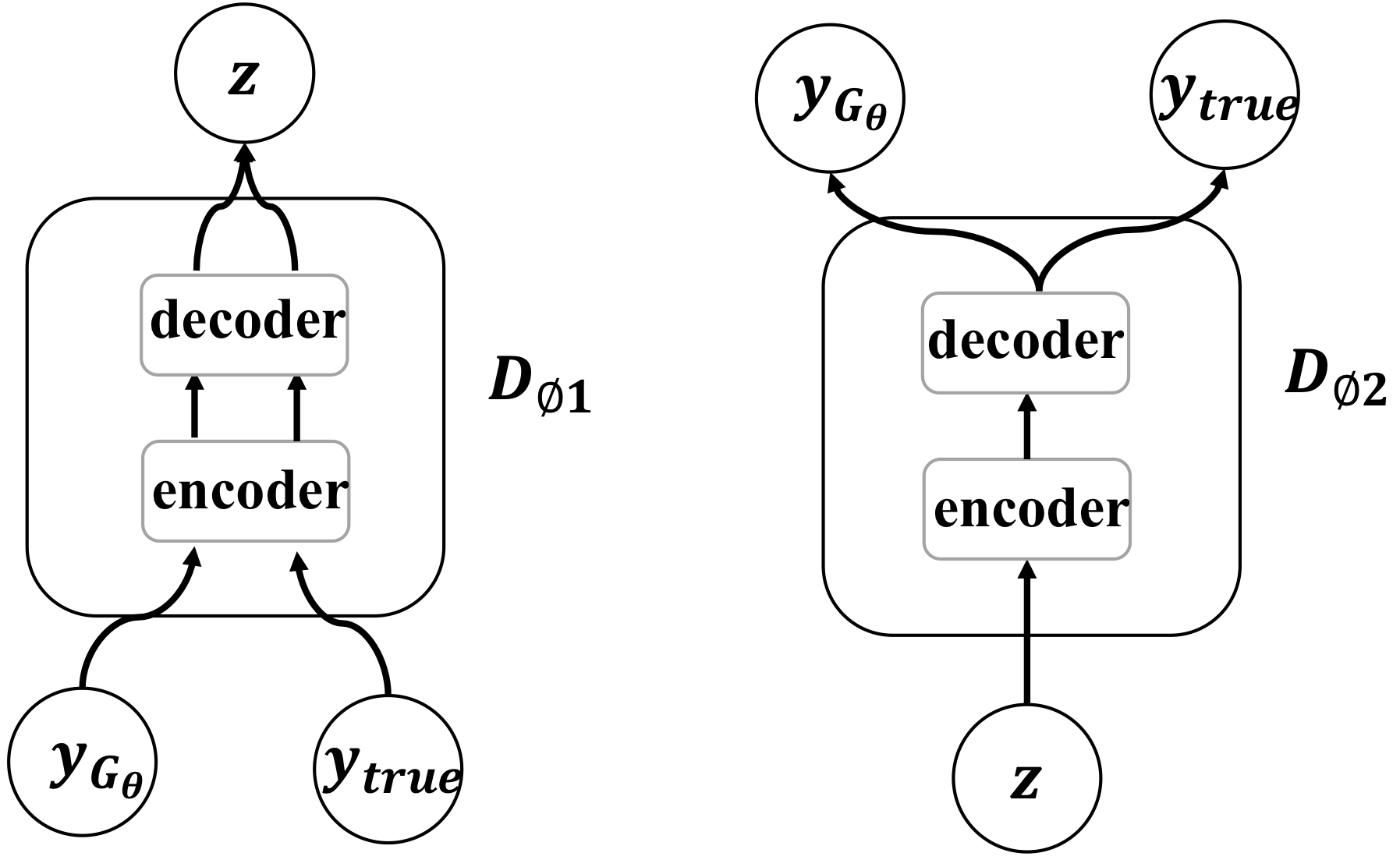} 
    \caption{Illustration of forward and backward generative discriminators. $y_{true}$ represents the real-world response. $y_{G_{\theta}}$ represents response generated by generator $G_{\theta}$. $z$ represents the true future turn. $D_{\phi 1}$ and $D_{\phi 2}$ represent forward and backward generative discriminator respectively. The forward generative discriminator produces the future turn $z$ based on the generated response $y_{G_{\theta}}$ or the real-world response $y_{true}$. The backward generative discriminator predicts the generated response $y_{G_{\theta}}$ or the real-world response $y_{true}$ based on the future turn $z$.}
    \label{fig:discriminator_architecture}
\end{figure}

As shown in Figure \ref{fig:discriminator_architecture}, in this paper, instead of modeling the joint probability $p(x,y)$, we introduce the future conversations $z$ and utilize the conditional probability $p(z|y)$ and $p(y|z)$ as rewards. The forward discriminator $p(z|y)$ outputs high rewards if $y$ is informative enough to perceive the subsequent future conversations , and a backward discriminator $p(y|z)$ encourages the generated response to be more coherent in terms of the following conversations and generates high rewards if the generated response bridge the gap between the query $x$ and the future conversation $z$. Note that, in order to induce the generated response to be informative and coherent in terms of both the given query and the future conversations, and stabilize the adversarial training process \cite{li2017adversarial,wu2018study}, we also optimize the generator by teacher forcing periodically.

\subsubsection{Forward Generative Discriminator}
Intuitively, in multi-turn conversations, a high-quality response not only responds to the query but also is informative enough to perceive future conversations. The forward generative discriminator takes in the response $y$ (the predicted response $y_{G_{\theta}}$ or the real-world response $y_{true}$), and generates the future turn $z$, a sequence of $K$ words. It discriminates whether $y$ is informative and appropriate enough to induce the future turn. 
 
In detail, for a response $y$ of $L$ words, the reward of generating the real-world future turn $z$ is defined as the averaged negative cross entropy of each word of $z$:
 \begin{equation}
  R_{1} \left( y \right) = \frac{1}{K} \sum_{k=1}^{K} \log D_{\phi 1}\left(z_{k} | y, z_{<k} \right).
  \label{equ:R1}
 \end{equation}
We maximize the reward for real-world response $y_{true}$ for generating the future response $z$ and minimize the reward for the generated response $y_{G_{\theta}}$ of predicting $z$. We expect the general, meaningless generated responses are of lower rewards while the informative responses are of higher rewards.
The loss function of the forward generative discriminators is formulated as follows:
 \begin{equation}
    \begin{array}{l}{J(\phi 1)=} \\ 
    {
    -\left( E_{ y_{true} \sim p_{\text{data}}} [R_{1}(y_{true})] - E_{ y_{G_{\theta}} \sim G_{\theta}} [R_{1}(y_{G_{\theta}})] \right)
    }
    \end{array}.
    \label{equ:D1}
\end{equation}

In contrast, the reward of the existing classifier-based discriminators are calculated as follows:
 $$
  R\left( y \right)=D_{\phi} \left( \operatorname{true} | x, y\right),
 $$
where $D_{\phi}$ is a binary classifier judging how likely $(x, y)$ is from the real-world data. One major problem of the classifier-based discriminator is that the reward is easy to saturate, where for a given context, different generated responses usually achieve similar rewards from the saturated region of the non-linear classification function like sigmoid \cite{xu2018diversity}. As a result, the discriminator fails to distinguish detailed fine-grained differences among the generated responses in such a situation. In forward generative discriminator, the response $y$ is differentiated by the ability of seeing the future few turns of conversations $z$. Such a cross-entropy based reward not only does not saturate but also discriminates the response in terms of $z$.

\subsubsection{Backward Generative Discriminator}
To further induce the generated response to be more coherent with both the preceding and the following conversations, we propose a backward generative discriminator $p(y|z)$.
 
Given the real-world future conversation $z$, the reward for $m^{th}$ word in response $y$ of $L$ word (the real-world response $y_{true}$, and the generated response $y_{G_{\theta}}$) is calculated at the word level:
  \begin{equation}
    R_{2} \left( y_{m} \right) = - \log D_{\phi 2} \left( y_{m} | z, y_{<m} \right).
    \label{equ:R2}
  \end{equation}
We maximize the reward $R_{2}$ for the real-world response $y_{true}$ and minimize the reward $R_{2}$ for the response $y_{G_{\theta}}$ produced by generator $G_{\theta}$. We formulate the loss function of the backward generative discriminator as follows:
  \begin{equation}
    \begin{array}{l}
    {J(\phi 2)=} \\ 
    {
    -\left( E_{y_{true} \sim p_{\text{data}}} [R_{2}(y_{true})] - E_{y_{G_{\theta}} \sim G_{\theta}} [R_{2}(y_{G_{\theta}})] \right)
    }
    \end{array}.
  \label{equ:D2}
  \end{equation}

If a response $y$ matches well with the given context $x$, but is irrelevant with the following conversations $z$, in previous discriminators, it may be endowed with a high reward. Whereas the backward discriminator models the generative probability $p(y|z)$ given the future conversation $z$, it induces the generated response to be more coherent by bridging the gap between the preceding and the subsequent conversations. We expect the responses which are coherent with both the preceding and the following conversations gain higher rewards while the responses that are irrelevant with the subsequent turns achieve lower rewards.

\subsection{Optimization}
In this work, the policy gradient method \cite{sutton2000policy,williams1992simple} is employed for optimization. The generator (policy) is trained to maximize the cumulative total reward of generated response:
  \begin{equation}
    J(\theta) = E_{y_{G_{\theta}} \sim G_{\theta}} \left(Q_{D_{\phi 1}, D_{\phi 2}}^{G_{\theta}} \left(x, y_{G_{\theta}}, z \right) \right),
  \label{equ:J1}
  \end{equation}
where $Q_{D_{\phi 1}, D_{\phi 2}}^{G_{\theta}}\left(x, y_{G_{\theta}}, z \right)$ is the cumulative total reward for a generated response $y_{G_{\theta}}$ starting from initial state $x$, taking action $a$ according to the policy $G_{\theta}$. The gradient of Eq.\eqref{equ:J1} is approximated using the likelihood ratio trick \cite{williams1992simple}:
  \begin{equation}
    \begin{array}{l}
    {\nabla_{\theta} J(\theta) \simeq} \\ 
    {
    \sum_{n=1}^{N} \sum_{m=1}^{L} R^{n}_{m} \nabla_{\theta} \log G_{\theta} \left(y^{n}_{m} | x, y^{n}_{<m}\right)
    }
    \end{array},
  \label{equ:J2}
  \end{equation}
where $N$ is the number of sampling via the policy $G_{\theta}$, and $R^{n}_{m}$ is the final reward of $m^{th}$ word in response $y^{n}_{G_{\theta}}$ by combining the reward $R1$ and $R2$ as $\sum_{i=m}^{L} \lambda^{i} \left( R1 \left( y^{n} \right) - MIN(R1) \right) R2\left(y^{n}_{i}\right)$. The term $\gamma$ is the discount rate. The $MIN(R1)$ is defined as the minimum response reward of each batch in training samples.

Only using policy gradient methods to optimize the generator directly will lead to a very fragile training process \cite{li2017adversarial}, because the generator never has access to the real-world response throughout the training process. Thus, we adopt the following three strategies to promote and stabilize the training process.\\
\textbf{Curriculum Learning Strategy.} For an utterance, the first $T$ words are optimized by MLE and the rest uses the policy gradient to calculate the loss. Then policy gradient is gradually adopted at every word \cite{li2016deep}.\\
\textbf{Baseline Strategy.} It facilitates the training process to be more steady by encouraging the model generates responses that achieve higher rewards than the baseline and suppressing the response generation with lower rewards compared with the baseline. In practice, we calculate the average word rewards of each batch in training samples as the baseline. When we only use the forward generative discriminator to judge the generated response, the baseline is set to the average response rewards of every batch in training samples.\\
\textbf{1-value reward Strategy.}
Following previous work \cite{li2016deep,li2017adversarial,xu2018diversity}, we also utilize teacher forcing to train the generator periodically. In this work, teacher forcing forces the generator to keep the given query in mind. We use the maximum likelihood estimation (MLE) objective in the teacher forcing phase, which can be viewed as setting the reward of the real-word response to 1 when using the policy gradient.

\section{Experiment}

\begin{table*}[thb]
\centering
\footnotesize 

\begin{tabular}{llllcccc}
\multicolumn{8}{c}{DailyDialog} \\ \hline
Models & Dist-1 & Dist-2 & Dist-3 & BLEU & Greedy & Average & Extrema \\ \hline
Seq2Seq-att & 0.0277 & 0.1625 & 0.3868 & 0.1878 & 0.4825 & 0.5993 & 0.3080 \\
Adver-REGS & 0.0541 & 0.2877 & 0.5542 & 0.2116 & 0.4857 & 0.6215 & 0.3542 \\
DP-GAN & 0.0656 & 0.3088 & 0.5630 & 0.1992 & 0.4749 & 0.6144 & 0.3409 \\ \hline
Posterior-GAN(F) & 0.0659 & 0.2995 & 0.5578 & 0.2067 & 0.4754 & 0.6218 & 0.3343 \\
Posterior-GAN(B) & 0.0578 & 0.2950 & 0.5545 & 0.2130 & 0.4818 & 0.6220 & 0.3442 \\
Posterior-GAN(A) & \bf 0.0678 & \bf 0.3549 & \bf 0.6006 & \bf 0.2183 & \bf 0.4916 & \bf 0.6260 & \bf 0.3544 \\ \hline
\multicolumn{8}{c}{OpenSubtitles (OSDb)} \\ \hline
Models & Dist-1 & Dist-2 & Dist-3 & BLEU & Greedy & Average & Extrema \\ \hline
Seq2Seq-att & 0.0016 & 0.0064 & 0.0150 & 0.1405 & 0.3900 & 0.4527 & 0.2243 \\
Adver-REGS & 0.0041 & 0.0136 & 0.0248 & 0.1609 & 0.4655 & 0.5523 & 0.2645 \\
DP-GAN & 0.0044 & 0.0143 & 0.0262 & 0.1484 & 0.4600 & 0.5509 & 0.2589 \\ \hline
Posterior-GAN(F) & 0.0049 & 0.0170 & 0.0322 & 0.1733 & 0.4753 & 0.5708 & 0.2573 \\
Posterior-GAN(B) & 0.0045 & 0.0148 & 0.0321 & 0.1756 & 0.4947 & 0.6217 & 0.2690 \\
Posterior-GAN(A) & \bf 0.0049 & \bf 0.0180 & \bf 0.0330 & \bf 0.1955 & \bf 0.4973 & \bf 0.6346 & \bf 0.2778 \\ \hline
\end{tabular}
\caption{The automatic metrics evaluation results. Higher is better. ``(F)'', ``(B)'' and ``(A)'' represent Posterior-GAN with a forward generative discriminator, a backward generative discriminator and both two discriminators, respectively.}
\label{experiment result: distinct/bleu/embedding-based metrics}
\end{table*}

\subsection{Datasets}

\textbf{DailyDialog:} This dataset consists of high-quality multi-turn dialog, which is provided by \citet{li2017dailydialog}. We construct the \textit{query-response-future turn} triples by treating each round in the dataset as response, three previous rounds as query, and three latter rounds as future turn. The length of response is limited to (5,40] by discarding the triples whose response is shorter than 5 words and truncating response over the maximum length to 40 words. The size of query and future turn is limited to less than 80 words. We randomly sample 28K, 3K, and 1.5K triples for training, validation, and testing sets, respectively.\\
\textbf{OpenSubtitles (OSDb):} OSDb\footnote{http://opus.lingfil.uu.se/OpenSubtitles.php} is a very large and noisy open-domain dataset containing roughly 60M-70M scripted lines. We first preprocess the triples as we do with DailyDialog, then select one subset in our experiment and split it into 1500K, 50K, and 25K triples for training, validation, and testing set, respectively.

\subsection{Comparison Models}

We compare the proposed Posterior-GAN with the following state-of-the-art models: \\
\textbf{Seq2Seq-att:} The generator is a sequence-to-sequence model \cite{sutskever2014sequence} with attention mechanism \cite{bahdanau2014neural}. A maximum likelihood estimation (MLE) objective is used to train the model.\\
\textbf{Adver-REGS:} Adver-REGS \cite{li2017adversarial} uses a sequence-to-sequence model to generate response. A binary classifier based discriminator calculates reward to train generator with policy gradient.\\
\textbf{DP-GAN:} DP-GAN \cite{xu2018diversity} also consists of a generator and a discriminator. Different from Adver-REGS, this discriminator is a cross-entropy based language model which alleviates the reward saturation problem.

\subsection{Training Details}

Based on the loss and the metrics on the validation set, we train the comparison models and our model with the following hyperparameters: The word embedding size is 256. The hidden size is set to 256. To conduct a fair comparison among all the models, We set the encoder layer to 2 and the decoder layer to 4. The encoder is a bidirectional LSTM. The vocabulary for DailyDialog and OpenSubtitles is of size 20,000 and 50,000, respectively. The batch size is set to 256 for pre-training and adversarial training. All the parameters is initialized using a normal distribution $\mathcal{N}(0,0.0001)$. All the models are trained end-to-end using Adam \cite{kingma2014adam} with a learning rate of 0.0001 and a global norm clipping at 2.0. For Adver-REGS, DP-GAN, and our model, before adversarial learning, we pre-train the generator for 10 epochs. In adversarial training, we alternatively train the generator every 1000 steps and optimize the discriminator every 5000 steps.

\subsection{Evaluation Metrics}

We evaluate the model in terms of following automatic evaluation metrics:
\begin{itemize}
    \item BLEU \cite{papineni2002bleu}, a word-overlapping based metric, which calculates word overlapping degree between the generated response and the real-world response. Recently plenty of work adopts its to reflect the lexical similarity of response \cite{li2016diversity,zhao2017learning,zhang2018reinforcing}.
    \item Embedding-based Metrics. Embedding Average (Average), Embedding Greedy (Greedy) and Embedding Extrema (Extrema) \cite{liu2016not} are used in the experiments. The three embedding-based metrics first calculate semantic embedding based on the vectors of all individual tokens in responses and then calculate the similarity between the generated response and the real-world response by cosine distance. They are widely used to evaluate the semantic similarity of response \cite{serban2017hierarchical,zhang2018generating,csaky2019improving}.
    \item Distinct. Dist-\{1,2,3\} are employed to reflect the degree of diversity of the generated responses, which are widely used in generative dialogue task \cite{li2016diversity,xu2018diversity,zhang2018generating}. The Dist-\{1,2,3\} represent the percentage (\%) of distinct unigrams/bigrams/trigrams.
\end{itemize}

\begin{table}[!t]
\centering
\footnotesize 

\begin{tabular}{l c c}
\multicolumn{3}{c}{DailyDialog} \\ 
\hline
Models & Coherence & Informativeness \\ 
\hline
Seq2Seq-att & 3.8550 & 3.8933 \\
Adver-REGS & 3.4717 & 3.3683 \\
DP-GAN & 3.4400 & 3.2350 \\ 
Posterior-GAN & \bf 3.2883 & \bf 3.2250 \\ 
\hline

\multicolumn{3}{c}{OpenSubtitles (OSDb)} \\ 
\hline
Models & Coherence & Informativeness \\ 
\hline
Seq2Seq-att & 3.8549 & 3.6952 \\
Adver-REGS & 4.0365 & 4.0432 \\
DP-GAN & 3.7638 & 3.8088 \\ 
Posterior-GAN & \bf 3.4806 & \bf 3.4567 \\ 
\hline
\end{tabular}

\caption{The human evaluation results. We calculate each score by averaging the rank of each model in corresponding metrics. Lower is better.}
\label{experiment result: human evaluation}
\end{table}

\subsection{Experimental Results}

\subsubsection{Overall Performance}
Table \ref{experiment result: distinct/bleu/embedding-based metrics} illustrates the evaluation results on lexical and semantic similarity metrics, and shows the diversity of the generated responses. Comparing Adver-REGS with DP-GAN, Adver-REGS performs better on BLEU and embedding-based similarities while DP-GAN generates more diverse responses in terms of Dist-\{1,2,3\}, which is consistent with the observation in \cite{xu2018diversity}. DP-GAN effectively improves the response diversity by utilizing the language model cross-entropy rewards while the performance on BLEU and embedding-based similarities do not witness similar improvements in our settings. Posterior-GAN achieves the best performance on all the automatic evaluation metrics on both corpora, indicating the superiority of the query-response-future turn triple training, enabled by the forward and backward generative discriminators, in comparison with the state-of-the-art generative approaches. And the improvements of our model are significant with \text{$p \leq 0.001$} (T-test).
 
\subsubsection{Ablation Test}
Comparing the forward and backward generative discriminator in Posterior-GAN, we observe that forward generative discriminator achieves better performance on the diversity metrics, whereas backward generative discriminator performs better on lexical and semantic similarities. The difference lies in that backward generative discriminator directly calculates the reward for all individual tokens of the generated response in terms of the future conversation, in the supplement of the generation perspective based on the query, and forward generative discriminator measures whether the generated response is informative enough to predict the subsequent real-world turns.

\subsubsection{Qualitative Evaluation}
Due to the known fact that quantitative metrics and human perception have a certain degree of deviation \cite{stent2005evaluating}, e.g., the conceptual difference of informativeness and diversity \cite{zhang2018generating}, we use human evaluation as a qualitative way to further evaluate our model and comparison models. We randomly select 200 samples from the test sets in the two dialogue datasets separately. Each sample consists of query, future turn and responses generated by different models. We invite three annotators to rank the generated responses with respect to two aspects: coherence and informativeness. Ties are allowed. Coherence indicates how likely the generated response is relevant to both query and future turn. Informativeness specifies how much the information related to the context is contained in the generated response. The annotators are all well educated and are ignorant of the models by which the responses are generated. To ensure a stable comparison, the evaluated models consist of 2 Seq2Sea-att models with different initialization, Adver-REGS, DP-GAN, and 3 Posterior-GAN outputs. The results of the human evaluation are shown in Table \ref{experiment result: human evaluation}. We also report the inter-annotator agreement to demonstrate the consistency of three annotators. The spearman's rank correlation coefficient for coherence and informativeness is 0.3948 and 0.3406, with $p < 0.0001$. Augmented with adversarial learning framework, DP-GAN, Adver-REGS, and Posterior-GAN all achieve better performance in comparison with the vanilla Seq2Seq-att model on DailyDialog, and DP-GAN performs better than Adver-REGS, similar to the observations in \cite{xu2018diversity}. Whereas, our model obtains substantial and consistent improvements in terms of coherence and informativeness on two public datasets, DailyDialog and Opensubtitles (OSDb).

\begin{table}[!t]
\centering
\footnotesize 

\begin{tabular}{p{8cm}}
\hline
\textbf{Query:} I need some flowers for my girlfriend. // No problem. Would you like some artificial carnations? // Oh, no. Carnations are not very elegant. Artificial flowers have no passion. \\
\hline
\textbf{Seq2Seq-att:} How how how much how much? I'd like.   \\
\textbf{Adver-REGS:} How did you want to spend? How much did.   \\
\textbf{DP-GAN:} How long did you want to spend? They was.   \\
\textbf{Posterior-GAN:} How much did you want to spend? It's very expensive.   \\
\hline
\textbf{Future turn:} Money is no object. // Our most elegant flower is golden lily. // I will take ten.  \\
\hline
\hline

\textbf{Query:} How may I help you today? // I need to open a second account. // What kind of account would you like to open?  \\
\hline
\textbf{Seq2Seq-att:} I want to need to deposit to deposit.  \\
\textbf{Adver-REGS:} I need to open the savings account.  \\
\textbf{DP-GAN:} I want to open cash at cash. i need  \\
\textbf{Posterior-GAN:} I need to transfer a savings account account. \\
\hline
\textbf{Future turn:} Do you have another account with us? // I sure do. // Would you like to transfer money from that account into your new one?   \\
\hline
\hline

\textbf{Query:} What's your case? // I was pulled over for running a red light, but I never did. // Do you believe that the officer lied? \\
\hline
\textbf{Seq2Seq-att:} Well, you was so in that. \\
\textbf{Adver-REGS:} We never like to lie. \\
\textbf{DP-GAN:} Because I've seen them. \\
\textbf{Posterior-GAN:} Of course, you can't know that. \\
\hline
\textbf{Future turn:} Your license plate was caught on camera? // A picture of my license plate was never taken. // Since there is no picture of your license plate on record, I'm going to let you go. \\
\hline
\end{tabular}

\caption{The responses generated by the proposed models and comparison models on DailyDialog.}
\label{experiment result: case study}
\end{table}

\subsection{Analysis: Why It Works}
In this section, we further analysis why future turn and two symmetric generative discriminators have a positive effect on the performance of model and increase informativeness and coherence of the response.

\subsubsection{Case Study}
We first show several examples in Table \ref{experiment result: case study}, which consist of query, future turn, and responses produced by different models on DailyDialog. It can be observed that the responses produced by our models are not only more consistent with the given query but also more coherent with future conversations. The comparison models are easy to generate repeated words, like ``how how how'', ``to deposit to deposit'', and hard to produce specific words related to future turn, which can reflect the ability to looking ahead. In the first example, it is more than clear that the responses generated by our models first respond to the query and then deepen the topic, which brings the conversation topic to continue the future turn. Similar observations also appear in other examples, but we do not show them for limited space.

\begin{table}[!t]
\centering
\footnotesize 

\begin{tabular}{@{}lcc@{}}
\hline
Models & \begin{tabular}[c]{@{}c@{}}Averaged Greedy Matching\\ for $y$ and ($x$/$z$) \end{tabular} & \begin{tabular}[c]{@{}c@{}}Frequency-based\\ Similarity\end{tabular} \\
\hline
Seq2Seq-att & 0.5942 & 0.5478 \\ 
Adver-REGS & 0.6403 & 0.7165 \\ 
DP-GAN & 0.6514 & 0.6739 \\ 
Posterior-GAN & \bf 0.7276 & \bf 0.7460 \\ 
\hline
\end{tabular}
\caption{The results of Embedding-based Averaged Greedy Matching for response $y$ with the given query $x$ and future conversations $z$, which reflects the coherence of the response, and Frequency-based Similarity, which illustrates the informativeness of the response.}
\label{experiment result: analyzation}
\end{table}

\subsubsection{Automatic Analysis}
To further demonstrate the above observations, we design two metrics to verify the superiority of the generated responses. The informativeness of the responses is reflected by comparing the word frequency similarity between the generated response and the ground-truth response, where the responses are represented as a vector, and each element in the vector is denoted as the frequency of a word. Here, we use 2350 most frequent words from the training set of DailyDialog corpus without stop words and meaningless words. To validate the coherence of the generated response, we calculate the average matching degree of the generated response with the given query and the subsequent real-world conversations by utilizing the embedding-based greedy matching metric, which prefers response with keywords that have high semantic similarity with those in the real-world context \cite{liu2016not}. The results are shown in Table \ref{experiment result: analyzation}.
Regarding both the frequency similarity and matching degree, our model consistently outperforms the comparison models, which indicates that our model generates more informative and coherent responses. 

\begin{figure*} 
    \centering 
    \includegraphics[width=.75\textwidth]{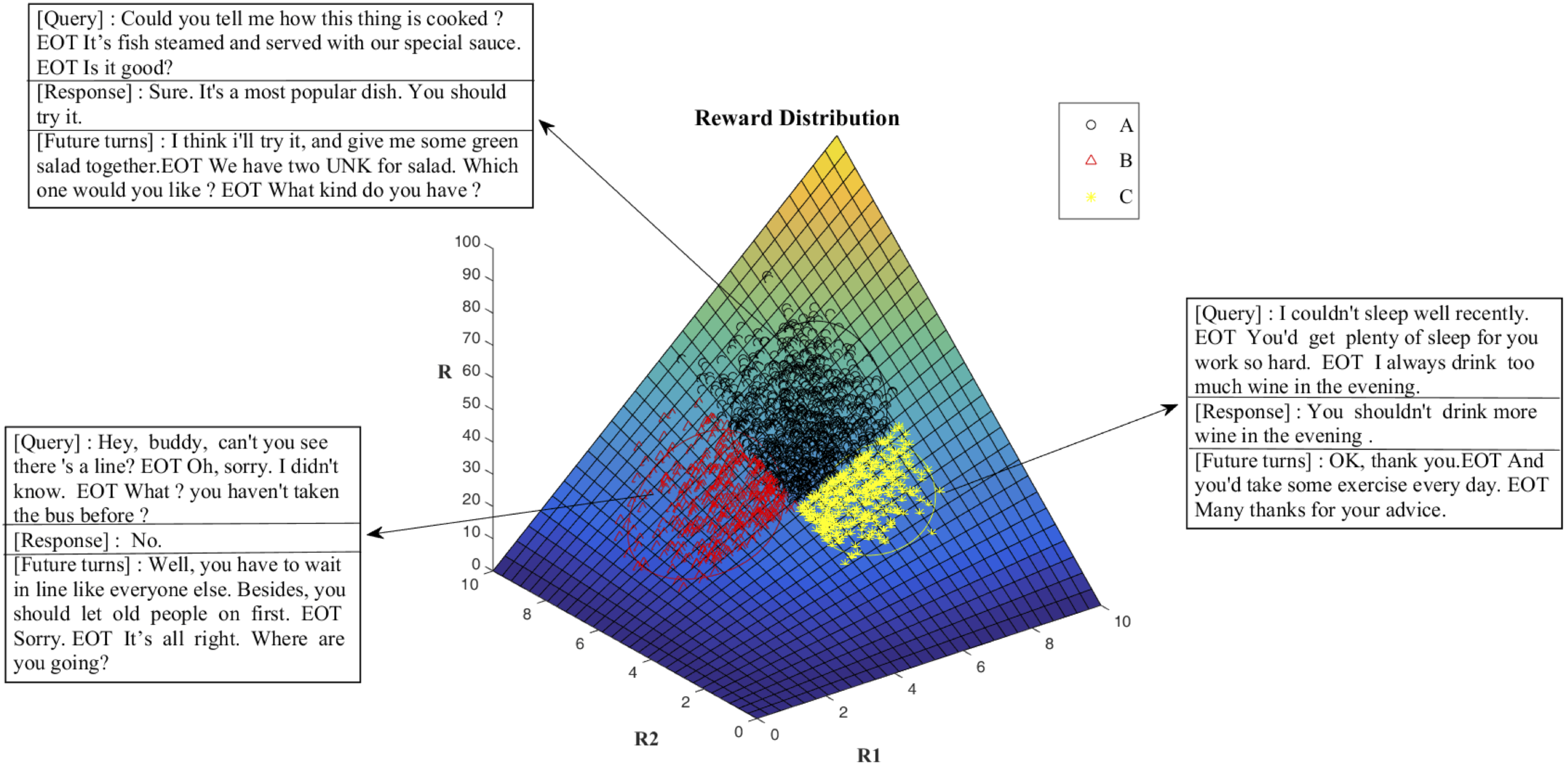} 
    \caption{The distribution of sample rewards calculated by the forward generative discriminator $R1$ and the backward generative discriminator $R2$ on DailyDialog. $R$ is the combination of $R1$ and $R2$. We use three regions A, B, and C to represent three types of samples. Samples in region A gain high rewards in both discriminators. Samples in region B achieve higher reward in the backward generative discriminator than in the forward one, while Samples in region C obtain higher reward in the forward generative discriminator than in the backward one.}
    \label{fig:rewards_distribution}
\end{figure*}

\subsubsection{Visualization}
We also visualize the reward distribution of two symmetric discriminators to get some insight into the behavior of the model on DailyDialog. Figure \ref{fig:rewards_distribution} illustrates the reward distribution and representative conversations. The reward distribution is roughly divided into three regions. We observe that some responses are of higher rewards on $R1$ (subtract the minimum value) but low rewards on $R2$ (the average $R2(y_{m})$ of response) as in the yellow region C of Figure \ref{fig:rewards_distribution}. The response from regions C provides specific information to answer the preceding context. The responses from red region B of Figure \ref{fig:rewards_distribution} are of higher rewards on $R2$, whereas the rewards on $R1$ is much lower. Although the response `no' is lack of informativeness, it is coherent to the given query and the future turn. By simultaneously integrating both $R1$ and $R2$, the generator pays more attention to responses that are not only informative but also coherent and achieves high rewards on both generative discriminators.

\section{Related Work}

A tremendous amount of effort has been paid to increase the informativeness and diversity of neural dialogue generation model. \citet{li2016diversity} adopted Maximum Mutual Information (MMI) as the objective function to decrease general response. \citet{li2016deep,zhang2018reinforcing} introduced reinforcement learning to facilitate the diversity of response with handcraft rewards. \citet{xing2017topic} incorporated topic information into the seq2eq based dialogue model to generate informative responses. \citet{cao2017latent,serban2017hierarchical,zhao2017learning,shen2018improving} applied CVAE to the seq2seq based dialogue model to increase utterance-level diversity and improve informativeness by generating a longer response. To enhance the coherence of the generated response, \citet{zhang2018reinforcing,xu2018better,csaky2019improving} manually designed an objective function that assesses the coherence of response with respect to the query. Whereas in our work, we handle the informativeness and coherence simultaneously by extending the conventional \textit{query-response} tuple learning into \textit{query-response-future turn} triple training. What is more, the proposed framework is optimized under the generative adversarial network instead of a handcraft learning objective.

Generative adversarial network \cite{goodfellow2014generative} has enjoyed certain success in dialogue response generation. \citet{li2017adversarial}  proposed adversarial training for dialogue generation. The model jointly trains two models, a generator (a Seq2Seq model) defining the probability of generating a dialogue sequence, and a discriminator labeling dialogues as human-generated or machine-generated. Since then, GAN based response generation models tended to solve the problem of repeated and “boring” expression such as GAN-AEL \cite{xu2017neural}, SeqGAN \cite{yu2017seqgan}, DP-GAN \cite{xu2018diversity}, MaskGAN \cite{fedus2018maskgan}, AIM \cite{zhang2018generating}, and DialogWAE \cite{gu2018dialogwae}. Our Posterior-GAN model differs from the above models in both the discriminator design and learning framework: DP-GAN uses a language-based discriminator to distinguish novel text from repeated text and assigns a low reward for repeated text and high reward for novel and fluent text; AIM exploits an embedding-based structured discriminator and uses Adversarial Information Maximization (AIM) model to generate informative and diverse responses; while in this paper, we propose a novel posterior adversarial learning framework to facilitate the \textit{query-response-future turn} modeling, where we adopt two encoder-decoder based generative discriminators, a forward and a backward discriminator. The two discriminators cooperatively discriminate the coherence and informativeness of the generated response, which bridges the gap between the preceding and the following conversations.

\section{Conclusion}

In this paper, we propose the \textit{query-response-future turn} triples instead of the conventional \textit{query-response} pairs for neural dialog response generation. To facilitate the triple modeling and alleviate the overproducing of generic and repetitive responses problem, Posterior-GAN that consists of a forward and a backward encoder-decoder based generative discriminator is further introduced. Augmented with future conversations and Posterior-GAN in training, detailed experiments and analysis demonstrate that the model effectively generates more informative and coherent responses.

\section{Acknowledgements}

This research is supported by National Key R\&D Program of China (No.2016YFB0801100), Beijing Natural Science Foundation (No.4172054, L181010), and National Basic Research Program of China (No.2013CB329605). Kan Li is the corresponding author.

\bibliography{aaai}
\bibliographystyle{aaai}

\end{document}